\pdfoutput=1

\documentclass[11pt]{article}

\usepackage{EMNLP2023}

\usepackage{times}
\usepackage{latexsym}

\usepackage[T1]{fontenc}

\usepackage[utf8]{inputenc}

\usepackage{microtype}

\usepackage{inconsolata}
\usepackage{amsfonts} 
\usepackage{graphicx}
\usepackage{caption}
\usepackage{xcolor}
\usepackage{subcaption}
\usepackage{multirow}
\usepackage{booktabs}
\usepackage{enumitem}
\usepackage{amsmath}
\usepackage{hyperref}
\usepackage{xspace}
\usepackage[capitalise]{cleveref}

\DeclareMathOperator*{\argmax}{arg\,max}

\newcommand{\sectionref}[1]{\S\ref{#1}}

\setlist[itemize]{itemsep=0pt, topsep=0pt, partopsep=0pt}

\newcommand{\method}{\texttt{SHARCS}\xspace}

\title{\method: Efficient Transformers through Routing \\ with Dynamic Width Sub-networks}

\author{Mohammadreza Salehi$^\dagger$ \enspace Sachin Mehta$^\dagger$ \enspace Aditya Kusupati$^\dagger$ \\ \textbf{Ali Farhadi$^{\dagger\diamond}$ \enspace Hannaneh Hajishirzi$^{\dagger\diamond}$} \vspace*{1mm}\\ $^\dagger$University of Washington \enspace $^\diamond$ Allen Institute for Artificial Intelligence \\ \texttt{\{mrsalehi,sacmehta,kusupati,ali,hannaneh\}@cs.washington.edu}}

\newcommand{\tablemodelsmain}{
\begin{table}[t!]
\tiny
\centering
\caption{\textbf{Comparison of different adaptive inference methods on MNLI-m dataset for different FLOP ranges}. For the full table, please see Appendix \ref{different_models_full_results}}
\vspace*{-2mm}
\resizebox{\columnwidth}{!}{
\begin{tabular}{cllllll}
\toprule[1.25pt]
\multirow{2}{*}{\begin{tabular}[c]{@{}c@{}}\textbf{FLOPs range}\\ \textbf{(Tera FLOPs)}\end{tabular}} & \multicolumn{1}{c}{} & \multicolumn{2}{c}{\textbf{Best Baseline}}                   & \multicolumn{1}{c}{} & \multicolumn{2}{c}{\textbf{\method}}                          \\ \cmidrule[0.75pt]{3-4} \cmidrule[0.75pt]{6-7} 
&\multicolumn{1}{c}{} & \multicolumn{1}{c}{\textbf{Acc.(\%) $\uparrow$}} & \multicolumn{1}{c}{\textbf{FLOPs $\downarrow$}} & \multicolumn{1}{c}{} & \multicolumn{1}{c}{\textbf{Acc.}} & \multicolumn{1}{c}{\textbf{FLOPs}} \\ \midrule[0.75pt]
\multicolumn{7}{c}{\textbf{Roberta (Acc: 87.6, FLOPs: 33.5)}}\\ \midrule[0.75pt]
0-10                                                                                &                      &  61.83                       &  9.56                         &                      & \textbf{76.37}                        & \textbf{4.91}                          \\
10-20                                                                               &                      & 84.19                        & 18                          &                      & \textbf{85.93}                        &  \textbf{17.92}                         \\
20-30                                                                               &                      & \textbf{87.53}                        & 29.86                          &                      &  87.38                       & \textbf{28.35}                          \\ \midrule[0.75pt]
\multicolumn{7}{c}{\textbf{BERT (Acc: 84.8, FLOPs: 33)}}                                                                                                                                                                                                                      \\ \midrule[0.75pt]
0-10                                                                                &                      &  48.87                       & \textbf{8.76}                          &                      &  \textbf{72.7}                       &  8.94                         \\
10-20                                                                               &                      &  72.50                       & 19.99                          &                      &  \textbf{81.61}                       & \textbf{16.73}                          \\
20-30                                                                               &                      &  83.27                       & 27.99                          &                      & \textbf{83.04}                        &  \textbf{22.76}                         \\ \midrule[0.75pt]
\multicolumn{7}{c}{\textbf{DistilBERT (Acc: 82.2, FLOPs: 16.57)}}                                                                                                                                                                                                                \\ \midrule[0.75pt]
0-5                                                                                &                      &  48.87                       & 4.7                          &                      &  \textbf{64.72}                       & \textbf{3.80}                          \\
5-10                                                                               &                      & 64.44                        & 9.95                          &                      &  \textbf{76.02}                       & \textbf{8.87}                          \\
10-15                                                                               &                      & 80.38                        & 14.91                          &                      &  \textbf{81.61}                       & \textbf{14.39}                          \\ \midrule[0.75pt]
\multicolumn{7}{c}{\textbf{DynaBERT 0.25 width (Acc: 83.9, FLOPs: 8.26)}}\\ \midrule[0.75pt]
0-4                                                                                &                      &  65.61                       & \textbf{3.78}                          &                      &  \textbf{78.92}                       & 3.99                          \\
4-6                                                                               &                      & 76.01                        & 5.42                          &                      & \textbf{81.48}                        & \textbf{5.21}                          \\
6-8                                                                               &                      & \textbf{83.86}                        & 7.92                          &                      & 83.40                        &  \textbf{7.33}                         \\ \bottomrule[1.25pt]
\end{tabular}
}
\vspace*{-2mm}
\label{table_models_main}
\end{table}
}

\newcommand{\tablemodelsappendix}{
\begin{table*}[ht!]
\centering
\tiny
\caption{Comparison of different adaptive inference methods on MNLI-m dataset for different FLOP ranges.}
\begin{tabular}{ccccccccccccccccccc} \toprule
\multirow{2}{*}{\begin{tabular}[c]{@{}c@{}}\textbf{FLOPs range}\\ \textbf{(Tera Flops)}\end{tabular}} & \multicolumn{1}{l}{} & \multicolumn{2}{c}{\textbf{RTJ}} & \multicolumn{1}{l}{} & \multicolumn{2}{c}{\textbf{DeeBERT}} & \multicolumn{1}{l}{} & \multicolumn{2}{c}{\textbf{PABEE}} & \multicolumn{1}{l}{} & \multicolumn{2}{l}{\textbf{FastBERT}}                         & \multicolumn{1}{l}{} & \multicolumn{2}{c}{\textbf{BERxiT}}                           & \multicolumn{1}{l}{} & \multicolumn{2}{c}{\textbf{\method}} \\ \cmidrule{3-4} \cmidrule{6-7} \cmidrule{9-10} \cmidrule{12-13} \cmidrule{15-16} \cmidrule{18-19} 
&\multicolumn{1}{l}{} & \textbf{Acc. $\uparrow$}       & \textbf{FLOPs $\downarrow$}      & \multicolumn{1}{l}{} & \textbf{Acc.}         & \textbf{FLOPs}        & \multicolumn{1}{l}{} & \textbf{Acc.}        & \textbf{FLOPs}       & \multicolumn{1}{l}{} & \multicolumn{1}{l}{\textbf{Acc.}} & \multicolumn{1}{l}{\textbf{FLOPs}} & \multicolumn{1}{l}{} & \multicolumn{1}{l}{\textbf{Acc.}} & \multicolumn{1}{l}{\textbf{FLOPs}} & \multicolumn{1}{l}{} & \textbf{Acc.}        & \textbf{FLOPs}      \\ \midrule
\multicolumn{19}{c}{\textbf{Roberta (Acc: 87.6, FLOPs: 33.5)}}                                                                                                                                                                                                                                                                                                                                                                                                                               \\ \midrule
0-10                                                                                &                      & 54.27      & 9.59       &                      & 61.83        & 9.56         &                      & 53.98       & 8.53        &                      & 45.39                    & 2.84                      &                      & 53.69                    & 8.41                      &                      & \textbf{76.37}           & \textbf{4.91}      \\
10-20                                                                               &                      & 82.29      & 19.77      &                      & 77.18        & 19.80        &                      & 84.19       & 18          &                      & 59.79                    & 15.06                     &                      & 83.39                    & 19.91                     &                      & \textbf{85.93}           & \textbf{17.92}      \\
20-30                                                                               &                      & 86.92      & 25.76      &                      & 81.35            & 23.67            &                      & 87.53       & 29.86       &                      & 77.24                    & 29.95                     &                      & 87.23                    & 27.92                     &                      & \textbf{87.38}           & \textbf{28.35}      \\ \midrule
\multicolumn{19}{c}{\textbf{BERT (Acc: 84.8, FLOPs: 33)}}                                                                                                                                                                                                                                                                                                                                                                                                                                  \\ \midrule
0-10                                                                                &                      & 47.32          & 9.46          &                      & 42.32        & 9.97         &                      & 42.97           & 7.51           &                      & 48.87                         & 8.76                         &                      & 46.86                    & 9.88                      &                      & \textbf{72.7}           & \textbf{8.94}         \\
10-20                                                                               &                      & 58.12          & 15.61          &                      & 61.57        & 19.89        &                      & 66.33           & 17.29           &                      & 70.25                        & 19.18                         &                      & 72.50                    & 19.99                     &                      & \textbf{81.61}           & \textbf{16.73}         \\
20-30                                                                               &                      & 78.96          & 28.53          &                      & 80.57        & 29.34        &                      & 81.13           &        27.66    &                      & 83.27                         & 27.99                         &                      & 80.82                    & 25.73                     &                      & \textbf{83.04}           & \textbf{22.76}         \\ \midrule
\multicolumn{19}{c}{\textbf{DistilBERT (Acc: 82.2, FLOPs: 16.57)}}                                                                                                                                                                                                                                                                                                                                                                                                                            \\ \midrule
0-5                                                                                 &                      & 41.92      & 2.76       &                      & 43.98           & 2.76           &                      & 41.45       & 2.76        &                      & 48.87                       & 4.7                        &                      & 41.47                    & 4.07                      &                      & \textbf{64.72}           & \textbf{3.80}         \\
5-10                                                                                &                      & 52.72      & 9.56       &                      & 59.87           & 9.45           &                      & 64.44       & 9.95        &                      & 64.24                       & 9.76                        &                      & 55.10                       & 9.76                        &                      & \textbf{76.02}           & \textbf{8.87}         \\
10-15                                                                               &                      & 72.93      & 13.90      &                      & 80.38           & 14.91           &                      & 79.29       & 14.12       &                      & 79.72                       & 14.84                        &                      & 60.25                    & 11.68                     &                      & \textbf{81.61}           & \textbf{14.39}         \\ \midrule
\multicolumn{19}{c}{\textbf{DynaBERT 0.25 width (Acc: 83.9, FLOPs: 8.26)}}                                                                                                                                                                                                                                                                                                                                                                          \\ \midrule
0-4                                                                                 &                      & 60.95          & 3.98          &                      & 56.13            & 3.38            &                      & 56.09          & 3.97          &                      & 65.61                       & 3.78                        &                      & 56.19                       & 3.58                        &                      & \textbf{78.92}          & \textbf{3.99}         \\
4-6                                                                                 &                      & 74.55          & 5.82          &                      & 75.26            & 5.62            &                      & 69.81          & 5.93          &                      & 76.01                       & 5.42                        &                      & -                       & -                        &                      & \textbf{81.48}           & \textbf{5.21}         \\
6-8                                                                                 &                      & 83.43          & 7.70          &                      & 83.36            & 7.70            &                      & 81.31          & 7.87          &                      & 83.86                       & 7.92                        &                      & 83.08                       & 7.60                        &                      & \textbf{83.40}          & \textbf{7.33}         \\ \bottomrule
\end{tabular}
\label{full_table_different_models}
\end{table*}
}

\newcommand{\tablespeed}{
\begin{table}[ht!]
\centering
\footnotesize
    \caption{\textbf{Inference speed up results on the QQP dataset on CPU for various adaptive inference techniques applied to BERT}. For \method we place the router after layers 2 and 4 and report two numbers.}\vspace*{-2mm}
    \resizebox{\columnwidth}{!}{
    \begin{tabular}{lcc} \toprule[1.5pt]
               & \textbf{Accuracy (\%) $\uparrow$}             & \textbf{Speed up ($\times$) $\uparrow$}           \\ \midrule[1pt]
    BERT & 90.90                 & 1                    \\ \midrule[1pt]
    DistilBERT & 88.50                 & 2                    \\
    BERxiT     & 85.60                & 2.04                 \\
    FastBERT   & 83.99                & 2.17                 \\
    DeeBERT    & 83.82                & 2.14                 \\
    PABEE      & 89.09                     &  2.06                    \\ 
    RTJ        &  84.44                    & 2                     \\ \midrule[1pt]
    \textbf{\method (layer 2)}      & 88.43                   &  \textbf{2.57} \\
    \textbf{\method (layer 4)}      & \textbf{90.05}                   & 2.05
    \\ \bottomrule[1.25pt]
    \end{tabular}}
    \label{table_speed_up}
    \vspace*{-3mm}
\end{table}
}

\newcommand{\tableablate}{
\begin{table}[h!]
\centering
\footnotesize
\caption{While using adaptive width with RoBERTa\textsubscript{base} on MNLI-m, \method router outperforms BERxiT router.}\vspace*{-3mm}
\footnotesize
\begin{tabular}{cc}
\toprule[1.5pt]
\textbf{Router on RoBERTa\textsubscript{base}}                      & \textbf{AUC} $\uparrow$ \\ \midrule[1pt]
\method    &  \textbf{0.78}               \\
BERxiT &  0.73                \\ \bottomrule[1.25pt]
\end{tabular}
\label{table_ablate}
\vspace*{-5pt}
\end{table}
}

\begin{document}
\maketitle
\begin{abstract}
We introduce \method for adaptive inference that takes into account the hardness of input samples. \method can train a router on any transformer network, enabling the model to direct different samples to sub-networks with varying widths. Our experiments demonstrate that: (1) \method outperforms or complements existing per-sample adaptive inference methods across various classification tasks in terms of accuracy vs. FLOPs; (2) \method generalizes across different architectures and can be even applied to compressed and efficient transformer encoders to further improve their efficiency; (3) \method can provide a $2\times$ inference speed up at an insignificant drop in accuracy. \end{abstract}

\section{Introduction}
\label{sec_intro}
Web-scale pretrained models, including Large Language Models (LLMs), are widely used in various applications \cite{bert, roberta, gpt3}. However, their computational resource requirements can be problematic, especially in environments with limited resources. To address this issue, more efficient methods are needed, particularly those that can run on-device with efficient inference \cite{distilbert}. 

Several methods (e.g., knowledge distillation \cite{knowledge_distillation}, pruning \cite{block_pruning,cofi}, and quantization \cite{qbert}) have been proposed to improve the inference efficiency of transformer-based models. While these methods are promising, one drawback is that the resulting model is static. This raises concerns about whether the model is too complex for simple samples and not complex enough for more challenging ones. To tackle this problem, previous work have investigated sample-adaptive inference methods that use varying amount of compute to process different input samples \cite{shallow_deep}. Two predominant approaches exist in the field of sample adaptive inference: early-exiting and token dropping. The former incorporates internal classifiers into intermediate layers of the model.  Various techniques have been explored for early exiting, including using confidence scores or entropy of internal classifier predictions \cite{fastbert,deebert}, using a module that predicts whether a layer should exit early \cite{berxit}, or implementing a patience-based method to adjust these internal predictions \cite{pabee}. The other category of sample-adaptive inference methods, token dropping, enhances efficiency by progressively decreasing sequence length as the forward pass proceeds layer by layer \cite{powerbert,transkimmer}.

In this paper, we propose 
\textbf{S}ample \textbf{H}ardness \textbf{A}ware \textbf{R}outing based on \textbf{C}onfidence \textbf{S}cores (\method\footnote{pronounced sharks.}), which is a novel category within the efficient sample adaptive inference domain. Our approach introduces \emph{training a light-weight router}. The router dynamically assigns input samples, based on their hardness \cite{vinformation}, to one of the \emph{sub-networks with varying widths}.  
Due to the lack of ground truth notion of hardness, we estimate sample hardness heuristically based on the network's prediction history during training for each sample. These estimates are then used as labels to train the router.

\vspace{2mm}
{We make the following contributions:}
\vspace{-2mm}
\begin{enumerate}[leftmargin=*]
    \setlength\itemsep{0pt}
    \item \method introduces a router that predicts the hardness of a sample and can be trained on \textit{any} transformer network. It enables dynamic input processing using sub-networks of varying widths, determined by the predicted hardness.
    \item \method delivers substantial efficiency improvements in terms of accuracy vs. FLOPs trade-off across different datasets \& transformer encoders. For example, on QQP \cite{qqp}, it reduces the FLOPs of RoBERTa\textsubscript{base} by  $2.75\times$ with only $1\%$  accuracy drop. Compared to other sample adaptive inference techniques, \method either outperforms them or can be paired with them to achieve even greater efficiency.
    \item The gains from \method can be realized in real-world deployment with significant latency reduction in CPU-based inference. On QQP, \method can speed up BERT\textsubscript{base} more than $2\times$  with less than 1\% drop in accuracy.
\end{enumerate}

\begin{figure}[t!]
     \centering
     \begin{subfigure}{0.38\textwidth}
         \centering
         \includegraphics[width=\textwidth]{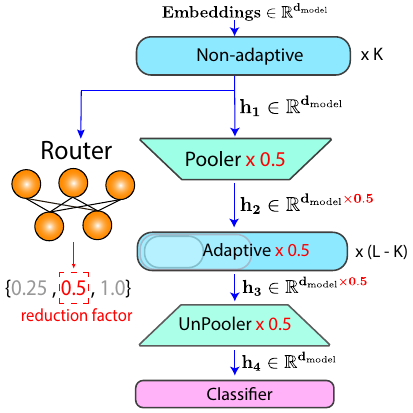}
         \caption{Inference.}
         \label{teaser_inference}
     \end{subfigure}\vspace{1mm}
     \begin{subfigure}{0.4\textwidth}
         \centering
         \includegraphics[width=\textwidth]{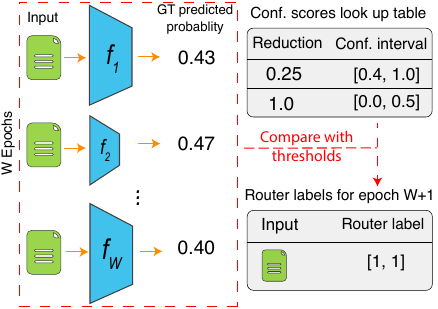}
         \caption{Generating hardness labels to train the router.}
         \label{teaser_train}
     \end{subfigure}
     \hfill
    \caption{\textbf{(a)} At inference the router selects the reduction factor. Red parts denote the changes enforced by the router. \textbf{(b)} Training the router. The confidence scores of the last $W$ epochs for each input sample is recorded. Then they are compared with confidence thresholds and labels for training the router are assigned accordingly.
    }
    \label{teaser}
\end{figure}




\section{Method}

We introduce \method for adaptive inference that takes into account the hardness of input samples. Our approach has three main steps: (1) Obtaining labels that represent sample hardness (\sectionref{section_hardness_label}); 
(2) Training a router that can predict sample hardness (\sectionref{subsection_learning_router}); 
(3) Adjusting the network's inference capacity according to the predicted hardness (\sectionref{subsection_reduce_width}).

\subsection{Estimating Hardness of a Sample}
\label{section_hardness_label}

Our objective is to learn a sample hardness aware router, enabling us to dynamically direct the input sample to one of the sub-networks with varying capacities (including the main network) for efficient inference. 
\begin{figure*}[ht!]
    \centering
    \includegraphics[width=0.87\textwidth]{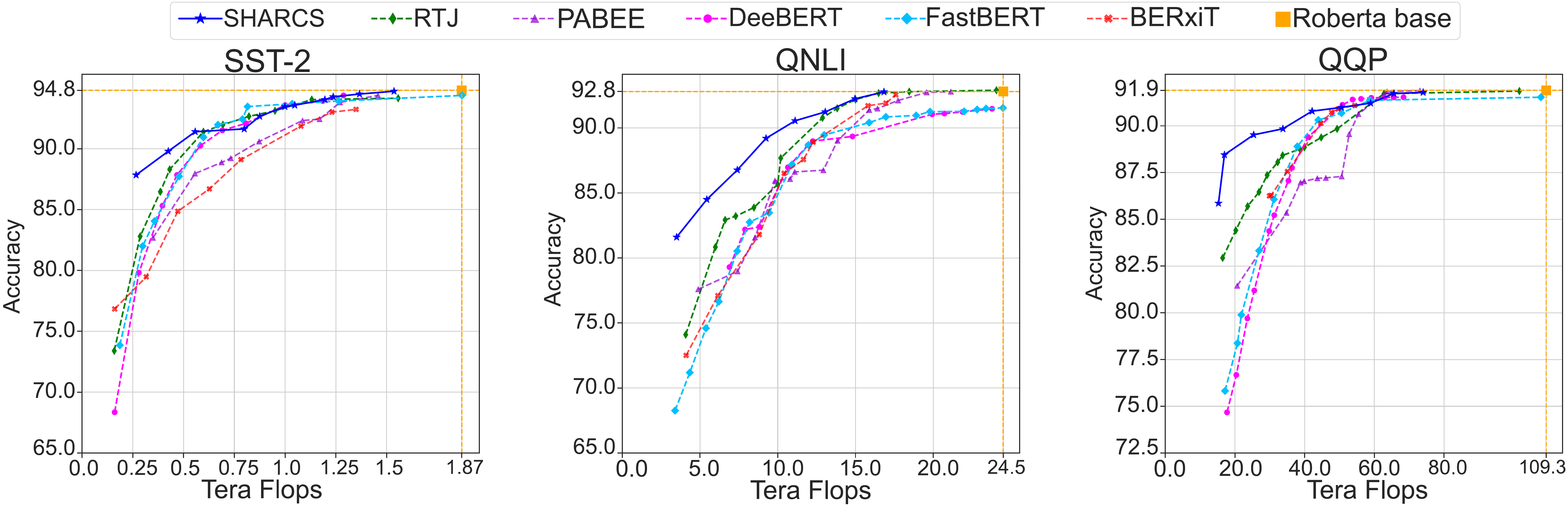}
    \caption{Results on the three of the sub-tasks in the GLUE benchmark. For the full set of plots on 8 tasks please refer to the appendix \ref{glue_full_results}. Best viewed in color.}
    \label{fig3:glue_results}
\end{figure*}
As there are no ground-truth labels that represent hardness, we  leverage  network's prediction history during training. 
We assume that there are $M$ possible hardness levels for a sample, with level $1$ being the easiest and $M$ being the hardest.
Our goal is to assign a hardness label $\hat{y} \in \{0, 1\}^M$ to samples in the training set and train the router with them. 
Toward this end, we employ a heuristic procedure which is illustrated in Figure \ref{teaser_train}: If the model predicts the ground truth class for a training sample with a probability within a confidence interval $[T_{low}^{i}, T_{high}^{i}]$, then the $i$th entry in the label $\hat{y}$ would be 1; otherwise it would be zero.
Here, $i$ ranges from 1 to $M$, and $T^i$'s are hyperparameters associated with hardness level $i$. 

Because of the stochastic nature of training, it is possible that the samples denoted as easy earlier will potentially be denoted as hard after that, posing instabilities while training the router. To mitigate such randomness, the $i$th entry is 1 only if the predicted ground truth probability is within the interval 
for a moving window of last $W$ epochs. The assigned hardness labels will be used as labels to train the router in the next epoch. We do not train the router during the first $W$ epochs and just train the network and record the network's predictions. Please see appendix \ref{appendix_history_window} for more details on the role of window size.

\subsection{Training Sample Hardness Aware Router}
\label{subsection_learning_router}
We split the main transformer network with $L$ layers into non-adaptive and adaptive parts, and incorporate the router between these networks (see Figure \ref{teaser_inference}). The non-adaptive network is comprised of the first $0\!<\!K<\!L\!-\!1$ layers while the adaptive part consists of the remaining $L-K$ layers. 
The adaptive component is a shared network that consists of sub-networks with varying widths, where the width of each sub-network is a fraction of the width of the network. 

More formally, sub-networks are associated with a given a set of reduction factors $\{r_i\}_{i=1}^M, 0 < r_i \leq 1$, where the width of $i$-th sub-network is $1/r_i\times$ smaller than that of main network.
We map the hardness level $i$ to the width reduction factor $r_i$. During training, for each input, we sample one of the reduction factors with entry 1 in the router label for that input and do the forward and backward pass with just the network associated with that reduction factor. 
During inference, given the output of non-adaptive network for input sample $x$, the objective of the router is to determine the width of sub-network in the adaptive module to process $x$. 
We train the sub-networks and the router with the following objective:

\begin{equation}
    \mathcal{L} = \lambda_{task} \cdot \mathcal{L}_{task} + \lambda_{router} \cdot \mathcal{L}_{router}
\end{equation}
where $\mathcal{L}_{router}$ is a binary cross-entropy loss between predicted and ground-truth hardness label, $\mathcal{L}_{task}$ is a task-specific loss, and $\lambda_{task}$ and $\lambda_{router}$ are loss weights which are hyper-parameters.

\label{sub_sample_hardness_aware_routing}

\subsection{Reducing Width of Adaptive Module}
\label{subsection_reduce_width}

The basic building block in both multi-head attention (MHA) module and feed forward network (FFN) in transformers is the linear layer \cite{transformer}. Given the fully-connected nature of linear layers, to reduce their width, we retain the leftmost $r \cdot d_{\text{model}}$ neurons in both input and output~\citep{kusupati2022matryoshka}. It is worth noting that as we reduce the input and output dimensions of matrix multiplications by a factor of $1/r$, the flops will be reduced by a factor of $1/r^2$. We follow a similar procedure for reducing the width of affine parameters in LayerNorm \cite{layernorm}. Also in our setup, we do not change the head dimensions in MHA and instead decrease the number of heads by a factor of $1/r$. As Figure \ref{teaser_inference} illustrates, we down-project the input hidden states to the adaptive layers using a pooler module and up-project it back before feeding to the single classifier for all of the sub-networks. 
Please see Appendix 
\ref{impl_details_appendix} for a detailed description of width reduction of different components in transformer layer.

\section{Experimental setup}

\paragraph{Datasets.} We evaluate \method on 8 classification tasks in the standard GLUE benchmark \citep{glue}: MNLI-m \cite{mnli}, QNLI \cite{glue,squad}, QQP \cite{qqp}, SST-2 \cite{sst}, MRPC \cite{mrpc}, RTE \cite{rte}, CoLA \cite{cola}, WNLI \cite{wnli}. 

\paragraph{Evaluation metrics.} Following previous work, we report the accuracy on the validation set, with an exception to CoLA for which we report Matthews correlation. We measure the total FLOPs on the validation set as it is invariant to the run time environment \cite{fastbert}.

\paragraph{Training details.} 
We train network with AdamW optimizer \cite{adamw}. We choose the number of epochs in $\{5, 10, 15\}$ and use learning rate in $\{2e-5$, $5e-5\}$ in our experiments. Please see  Appendix \ref{training_details} for more details.

\section{Results}
\subsection{\method is Better}
\label{results_different_datasets}
We compare \method with existing sample adaptive inference methods: RTJ \citep{RTJ}, DeeBERT \citep{deebert}, FastBERT \citep{fastbert}, BERxiT \citep{berxit}, and PABEE \citep{pabee}. 
Figure \ref{fig3:glue_results} shows the accuracy-FLOPs trade-off plots for different sample adaptive approaches with RoBERTa\textsubscript{base} model on three GLUE subtasks (See appendix \ref{glue_full_results} for the full set of plots). 
Our results show that \method significantly outperforms other methods, especially in the low-FLOPs regime. This can suggest that by substantially reducing the width of deeper layers in the model, \method can achieve a significant reduction in FLOPs. Furthermore, we can maintain the accuracy better compared to fully skipping deeper layers as commonly done in early exiting methods.

\tablemodelsmain

\subsection{\method is Model Agnostic}
\method can be seamlessly integrated with any state-of-the-art non-efficient and efficient transformer-based encoders to further improve their efficiency. 
We use \method and the baseline sample adaptive inference methods in (\sectionref{results_different_datasets}) with two standard non-efficient  (RoBERTa\textsubscript{base} and BERT-Base\textsubscript{base}) and efficient (DistilBERT \cite{distilbert} and DynaBERT \cite{dynabert}) models. Table \ref{table_models_main} shows that \method consistently improves the efficiency of different models while maintaining the accuracy better than other baseline sample adaptive methods. Note that, for brevity, we have selected the highest accuracy among all the sample adaptive baselines for each FLOPs range; please see Table \ref{full_table_different_models} in Appendix \ref{different_models_full_results} for the full set of results.
As an example, \method can reduce the FLOPs of BERT\textsubscript{base} to half with near 3\% reduction in accuracy whereas in other methods the drop is more than 12\%. Interestingly, \method can further improve the efficiency of already optimized networks. For instance, inference with \method on DynaBERT 0.25 width takes 10-15\% less FLOPs with less than 1\% drop in accuracy. More accuracy vs. FLOPs trade-off results for DynaBERT 0.25 and DistilBERT can be found in Figure \ref{fig:dyna_acc_flops_plot} and \ref{fig:distilbert_acc_flops_plot}.

\subsection{\method is Fast}
We compare the latency of \method applied to BERT\textsubscript{base} with the baseline adaptive approaches on the QQP \cite{qqp} dataset. The latency measurements are conducted on an Intel Xeon Gold 6230 CPU with a batch size of 1. We use two reduction factors $\{0.25, 1.0\}$ and place the router after layer 2 or 4 to get different accuracy-speed up trade offs. We keep the speed up between 2 to 3 and report the best accuracy for each method in this range. Table \ref{table_speed_up} shows that our method obtains higher or comparable accuracy in this speed up range. Interestingly, \method achieves a performance on par with DistilBERT, which is trained with distillation on a much larger dataset, but with higher speed up. 

\subsection{\method Paired with Token Dropping}
As discussed in \cref{sec_intro}, in addition to early-exiting, token dropping is another well-known approach to sample-adaptive inference efficiency \cite{powerbert,transkimmer}. Token dropping enhances efficiency by progressively dropping tokens layer by layer, thereby decreasing the hidden states sequence length as the forward pass proceeds. In contrast, \method improves efficiency via reducing the transformer block's width. Therefore token dropping and \method should not interfere with each other and in fact, they can be paired together to bring more efficiency. To showcase this, we combine \method with a RoBERTa\textsubscript{base} network that has already been made more efficient using Transkimmer token dropping. Table \ref{sharcs_transkimer} shows that on QNLI dataset, \method reduces the FLOPs of RoBERTa + Transkimmer by 40\% with a negligible drop in accuracy.

\tablespeed

\begin{table}[h!]
    \caption{\textbf{\method can be applied to RoBERTa in addition to Transkimmer to bring more efficiency.} The decrease in FLOPs and Accuracy in parantheses are with respect to RoBERTa + Transkimmer.}
    \resizebox{\columnwidth}{!}{
        \begin{tabular}{@{}lcc@{}}
            \toprule[1.5pt]
             & \textbf{QNLI Acc.} & \textbf{TeraFLOPs} \\ \midrule[1pt]
            RoBERTa Base & 92.8\% & 24.5 \\
            +Transkimmer & 89.45\% & 14.08 \\ \midrule[1pt]
            \textbf{+Transkimmer + \method} & 88.83\% (-0.62\%) & \textbf{8.56 ($\sim$40\%$\downarrow$)} \\ \bottomrule[1.25pt]
        \end{tabular}
    }
    \label{sharcs_transkimer}
\end{table}

\subsection{Ablating the Router}
To show the efficacy of the router and its training strategy in \method, we replace it with the routing strategy used in BERxiT \cite{berxit}: while training two sub-networks with reduction factors $\{0.25, 1.0\}$, we feed the training sample to both of them; if a sub-network correctly classifies the sample, the router label for that sub-network in that iteration would be 1. Otherwise, it would be zero. The backward pass is then done with all the losses of sub-networks and the router loss. Table \ref{table_ablate} shows the area under the normalized accuracy FLOPs curve (AUC) for both methods on MNLI-m. Please find more detailed ablation results and experiments in Appendix \ref{appendix_ablation_discussion}.
\tableablate

\section{Conclusion}
We presented \method as a new sample adaptive inference approach that can improve any network's inference efficiency. \method incorporated a light-weight router which is trained with a novel approach using the confidence of the network predictions during training. Our experiments showed the superiority or complementary role of \method compared to other sample adaptive inference methods across various datasets and backbones.

\section*{Limitations}
While the router and training approach in \method are general purpose, a limitation of this paper is its focus solely on studying the impact on transformer encoders. Nonetheless, decoder-only \cite{gpt, gpt2} and encoder-decoder \cite{T5, bart} models are widely used classes of models for which we plan to integrate \method in the future work. It should be noted that although our approach can be applied to regression tasks with minor adjustments, this work does not include any results for regression tasks.

\section*{Acknowledgements}
We thank members of RAIVN and H2Lab for helpful discussion and feedback, and Hyak cluster team at the University of Washington for infrastructure support. This research was supported by NSF IIS-2044660, ONR MURI N00014- 18-1-2670, and gifts from AI2, Google and Apple.

\bibliography{emnlp2023}

\begin{thebibliography}{46}
\expandafter\ifx\csname natexlab\endcsname\relax\def\natexlab#1{#1}\fi

\bibitem[{Ba et~al.(2016)Ba, Kiros, and Hinton}]{layernorm}
Jimmy~Lei Ba, Jamie~Ryan Kiros, and Geoffrey~E. Hinton. 2016.
\newblock \href {http://arxiv.org/abs/1607.06450} {Layer normalization}.

\bibitem[{Brown et~al.(2020)Brown, Mann, Ryder, Subbiah, Kaplan, Dhariwal,
  Neelakantan, Shyam, Sastry, Askell, Agarwal, Herbert{-}Voss, Krueger,
  Henighan, Child, Ramesh, Ziegler, Wu, Winter, Hesse, Chen, Sigler, Litwin,
  Gray, Chess, Clark, Berner, McCandlish, Radford, Sutskever, and
  Amodei}]{gpt3}
Tom~B. Brown, Benjamin Mann, Nick Ryder, Melanie Subbiah, Jared Kaplan,
  Prafulla Dhariwal, Arvind Neelakantan, Pranav Shyam, Girish Sastry, Amanda
  Askell, Sandhini Agarwal, Ariel Herbert{-}Voss, Gretchen Krueger, Tom
  Henighan, Rewon Child, Aditya Ramesh, Daniel~M. Ziegler, Jeffrey Wu, Clemens
  Winter, Christopher Hesse, Mark Chen, Eric Sigler, Mateusz Litwin, Scott
  Gray, Benjamin Chess, Jack Clark, Christopher Berner, Sam McCandlish, Alec
  Radford, Ilya Sutskever, and Dario Amodei. 2020.
\newblock \href {http://arxiv.org/abs/2005.14165} {Language models are few-shot
  learners}.
\newblock \emph{CoRR}, abs/2005.14165.

\bibitem[{Dagan et~al.(2005)Dagan, Glickman, and Magnini}]{rte}
Ido Dagan, Oren Glickman, and Bernardo Magnini. 2005.
\newblock \href {http://www.cs.biu.ac.il/~glikmao/rte05/} {The pascal
  recognising textual entailment challenge}.
\newblock In \emph{Proceedings of the PASCAL Challenges Workshop on Recognising
  Textual Entailment}.

\bibitem[{Devlin et~al.(2018)Devlin, Chang, Lee, and Toutanova}]{bert}
Jacob Devlin, Ming{-}Wei Chang, Kenton Lee, and Kristina Toutanova. 2018.
\newblock \href {http://arxiv.org/abs/1810.04805} {{BERT:} pre-training of deep
  bidirectional transformers for language understanding}.
\newblock \emph{CoRR}, abs/1810.04805.

\bibitem[{Devvrit et~al.(2023)Devvrit, Kudugunta, Kusupati, Dettmers, Chen,
  Dhillon, Tsvetkov, Hannaneh, Kakade, Farhadi, and
  Jain}]{devvrit2023matformer}
Devvrit, Sneha Kudugunta, Aditya Kusupati, Tim Dettmers, Kaifeng Chen, Inderjit
  Dhillon, Yulia Tsvetkov, Hajishirzi Hannaneh, Sham Kakade, Ali Farhadi, and
  Prateek Jain. 2023.
\newblock Matformer: Nested transformer for elastic inference.
\newblock \emph{arXiv preprint arxiv:2310.07707}.

\bibitem[{Dolan and Brockett(2005)}]{mrpc}
William~B. Dolan and Chris Brockett. 2005.
\newblock \href {https://aclanthology.org/I05-5002} {Automatically constructing
  a corpus of sentential paraphrases}.
\newblock In \emph{Proceedings of the Third International Workshop on
  Paraphrasing ({IWP}2005)}.

\bibitem[{Ethayarajh et~al.(2021)Ethayarajh, Choi, and
  Swayamdipta}]{vinformation}
Kawin Ethayarajh, Yejin Choi, and Swabha Swayamdipta. 2021.
\newblock \href {http://arxiv.org/abs/2110.08420} {Information-theoretic
  measures of dataset difficulty}.
\newblock \emph{CoRR}, abs/2110.08420.

\bibitem[{Goyal et~al.(2020)Goyal, Choudhury, Raje, Chakaravarthy, Sabharwal,
  and Verma}]{powerbert}
Saurabh Goyal, Anamitra~R. Choudhury, Saurabh~M. Raje, Venkatesan~T.
  Chakaravarthy, Yogish Sabharwal, and Ashish Verma. 2020.
\newblock \href {http://arxiv.org/abs/2001.08950} {Power-bert: Accelerating
  bert inference via progressive word-vector elimination}.

\bibitem[{Guan et~al.(2022)Guan, Li, Leng, Lin, and Guo}]{transkimmer}
Yue Guan, Zhengyi Li, Jingwen Leng, Zhouhan Lin, and Minyi Guo. 2022.
\newblock \href {https://doi.org/10.18653/v1/2022.acl-long.502} {Transkimmer:
  Transformer learns to layer-wise skim}.
\newblock In \emph{Proceedings of the 60th Annual Meeting of the Association
  for Computational Linguistics (Volume 1: Long Papers)}, pages 7275--7286,
  Dublin, Ireland. Association for Computational Linguistics.

\bibitem[{Hendrycks and Gimpel(2016)}]{gelu}
Dan Hendrycks and Kevin Gimpel. 2016.
\newblock \href {http://arxiv.org/abs/1606.08415} {Bridging nonlinearities and
  stochastic regularizers with gaussian error linear units}.
\newblock \emph{CoRR}, abs/1606.08415.

\bibitem[{Hinton et~al.(2015)Hinton, Vinyals, and
  Dean}]{knowledge_distillation}
Geoffrey Hinton, Oriol Vinyals, and Jeff Dean. 2015.
\newblock \href {http://arxiv.org/abs/1503.02531} {Distilling the knowledge in
  a neural network}.

\bibitem[{Hou et~al.(2020)Hou, Huang, Shang, Jiang, Chen, and Liu}]{dynabert}
Lu~Hou, Zhiqi Huang, Lifeng Shang, Xin Jiang, Xiao Chen, and Qun Liu. 2020.
\newblock \href
  {https://proceedings.neurips.cc/paper_files/paper/2020/file/6f5216f8d89b086c18298e043bfe48ed-Paper.pdf}
  {Dynabert: Dynamic bert with adaptive width and depth}.
\newblock In \emph{Advances in Neural Information Processing Systems},
  volume~33, pages 9782--9793. Curran Associates, Inc.

\bibitem[{Jiao et~al.(2019)Jiao, Yin, Shang, Jiang, Chen, Li, Wang, and
  Liu}]{tinybert}
Xiaoqi Jiao, Yichun Yin, Lifeng Shang, Xin Jiang, Xiao Chen, Linlin Li, Fang
  Wang, and Qun Liu. 2019.
\newblock \href {http://arxiv.org/abs/1909.10351} {Tinybert: Distilling {BERT}
  for natural language understanding}.
\newblock \emph{CoRR}, abs/1909.10351.

\bibitem[{Kaya and Dumitras(2018)}]{shallow_deep}
Yigitcan Kaya and Tudor Dumitras. 2018.
\newblock \href {http://arxiv.org/abs/1810.07052} {How to stop off-the-shelf
  deep neural networks from overthinking}.
\newblock \emph{CoRR}, abs/1810.07052.

\bibitem[{Kim et~al.(2021)Kim, Gholami, Yao, Mahoney, and Keutzer}]{ibert}
Sehoon Kim, Amir Gholami, Zhewei Yao, Michael~W Mahoney, and Kurt Keutzer.
  2021.
\newblock I-bert: Integer-only bert quantization.
\newblock \emph{International Conference on Machine Learning (Accepted)}.

\bibitem[{Kusupati et~al.(2022)Kusupati, Bhatt, Rege, Wallingford, Sinha,
  Ramanujan, Howard-Snyder, Chen, Kakade, Jain et~al.}]{kusupati2022matryoshka}
Aditya Kusupati, Gantavya Bhatt, Aniket Rege, Matthew Wallingford, Aditya
  Sinha, Vivek Ramanujan, William Howard-Snyder, Kaifeng Chen, Sham Kakade,
  Prateek Jain, et~al. 2022.
\newblock Matryoshka representation learning.
\newblock \emph{Advances in Neural Information Processing Systems},
  35:30233--30249.

\bibitem[{Lagunas et~al.(2021)Lagunas, Charlaix, Sanh, and
  Rush}]{block_pruning}
Fran{\c{c}}ois Lagunas, Ella Charlaix, Victor Sanh, and Alexander~M. Rush.
  2021.
\newblock \href {http://arxiv.org/abs/2109.04838} {Block pruning for faster
  transformers}.
\newblock \emph{CoRR}, abs/2109.04838.

\bibitem[{Levesque et~al.(2012)Levesque, Davis, and Morgenstern}]{wnli}
Hector~J. Levesque, Ernest Davis, and Leora Morgenstern. 2012.
\newblock \href {https://cs.nyu.edu/faculty/davise/papers/WSKR2012.pdf} {The
  {Winograd} {Schema} {Challenge}}.
\newblock In \emph{Proceedings of the {Thirteenth} {International} {Conference}
  on {Principles} of {Knowledge} {Representation} and {Reasoning}}, {KR}'12,
  pages 552--561. AAAI Press, Rome, Italy.

\bibitem[{Lewis et~al.(2020)Lewis, Liu, Goyal, Ghazvininejad, Mohamed, Levy,
  Stoyanov, and Zettlemoyer}]{bart}
Mike Lewis, Yinhan Liu, Naman Goyal, Marjan Ghazvininejad, Abdelrahman Mohamed,
  Omer Levy, Veselin Stoyanov, and Luke Zettlemoyer. 2020.
\newblock \href {https://doi.org/10.18653/v1/2020.acl-main.703} {{BART}:
  Denoising sequence-to-sequence pre-training for natural language generation,
  translation, and comprehension}.
\newblock In \emph{Proceedings of the 58th Annual Meeting of the Association
  for Computational Linguistics}, pages 7871--7880, Online. Association for
  Computational Linguistics.

\bibitem[{Liu et~al.(2020)Liu, Zhou, Wang, Zhao, Deng, and Ju}]{fastbert}
Weijie Liu, Peng Zhou, Zhiruo Wang, Zhe Zhao, Haotang Deng, and Qi~Ju. 2020.
\newblock \href {https://doi.org/10.18653/v1/2020.acl-main.537} {{F}ast{BERT}:
  a self-distilling {BERT} with adaptive inference time}.
\newblock In \emph{Proceedings of the 58th Annual Meeting of the Association
  for Computational Linguistics}, pages 6035--6044, Online. Association for
  Computational Linguistics.

\bibitem[{Liu et~al.(2019{\natexlab{a}})Liu, Ott, Goyal, Du, Joshi, Chen, Levy,
  Lewis, Zettlemoyer, and Stoyanov}]{roberta}
Yinhan Liu, Myle Ott, Naman Goyal, Jingfei Du, Mandar Joshi, Danqi Chen, Omer
  Levy, Mike Lewis, Luke Zettlemoyer, and Veselin Stoyanov. 2019{\natexlab{a}}.
\newblock \href {http://arxiv.org/abs/1907.11692} {Roberta: {A} robustly
  optimized {BERT} pretraining approach}.
\newblock \emph{CoRR}, abs/1907.11692.

\bibitem[{Liu et~al.(2019{\natexlab{b}})Liu, Mu, Zhang, Guo, Yang, Cheng, and
  Sun}]{metapruning}
Zechun Liu, Haoyuan Mu, Xiangyu Zhang, Zichao Guo, Xin Yang, Kwang{-}Ting~(Tim)
  Cheng, and Jian Sun. 2019{\natexlab{b}}.
\newblock \href {http://arxiv.org/abs/1903.10258} {Metapruning: Meta learning
  for automatic neural network channel pruning}.
\newblock \emph{CoRR}, abs/1903.10258.

\bibitem[{Loshchilov and Hutter(2019)}]{adamw}
Ilya Loshchilov and Frank Hutter. 2019.
\newblock \href {http://arxiv.org/abs/1711.05101} {Decoupled weight decay
  regularization}.

\bibitem[{Michel et~al.(2019)Michel, Levy, and Neubig}]{remove_heads}
Paul Michel, Omer Levy, and Graham Neubig. 2019.
\newblock \href
  {https://proceedings.neurips.cc/paper_files/paper/2019/file/2c601ad9d2ff9bc8b282670cdd54f69f-Paper.pdf}
  {Are sixteen heads really better than one?}
\newblock In \emph{Advances in Neural Information Processing Systems},
  volume~32. Curran Associates, Inc.

\bibitem[{Radford et~al.(2018)Radford, Narasimhan, Salimans, and
  Sutskever}]{gpt}
Alec Radford, Karthik Narasimhan, Tim Salimans, and Ilya Sutskever. 2018.
\newblock Improving language understanding by generative pre-training.

\bibitem[{Radford et~al.(2019)Radford, Wu, Child, Luan, Amodei, and
  Sutskever}]{gpt2}
Alec Radford, Jeff Wu, Rewon Child, David Luan, Dario Amodei, and Ilya
  Sutskever. 2019.
\newblock Language models are unsupervised multitask learners.

\bibitem[{Raffel et~al.(2019)Raffel, Shazeer, Roberts, Lee, Narang, Matena,
  Zhou, Li, and Liu}]{T5}
Colin Raffel, Noam Shazeer, Adam Roberts, Katherine Lee, Sharan Narang, Michael
  Matena, Yanqi Zhou, Wei Li, and Peter~J. Liu. 2019.
\newblock \href {http://arxiv.org/abs/1910.10683} {Exploring the limits of
  transfer learning with a unified text-to-text transformer}.
\newblock \emph{CoRR}, abs/1910.10683.

\bibitem[{Rajpurkar et~al.(2016)Rajpurkar, Zhang, Lopyrev, and Liang}]{squad}
Pranav Rajpurkar, Jian Zhang, Konstantin Lopyrev, and Percy Liang. 2016.
\newblock \href {https://doi.org/10.18653/v1/D16-1264} {{SQ}u{AD}: 100,000+
  questions for machine comprehension of text}.
\newblock In \emph{Proceedings of the 2016 Conference on Empirical Methods in
  Natural Language Processing}, pages 2383--2392, Austin, Texas. Association
  for Computational Linguistics.

\bibitem[{Sanh et~al.(2019)Sanh, Debut, Chaumond, and Wolf}]{distilbert}
Victor Sanh, Lysandre Debut, Julien Chaumond, and Thomas Wolf. 2019.
\newblock \href {http://arxiv.org/abs/1910.01108} {Distilbert, a distilled
  version of {BERT:} smaller, faster, cheaper and lighter}.
\newblock \emph{CoRR}, abs/1910.01108.

\bibitem[{Sanh et~al.(2020)Sanh, Wolf, and Rush}]{movement_pruning}
Victor Sanh, Thomas Wolf, and Alexander~M. Rush. 2020.
\newblock \href {http://arxiv.org/abs/2005.07683} {Movement pruning: Adaptive
  sparsity by fine-tuning}.
\newblock \emph{CoRR}, abs/2005.07683.

\bibitem[{Schwartz et~al.(2020)Schwartz, Stanovsky, Swayamdipta, Dodge, and
  Smith}]{RTJ}
Roy Schwartz, Gabriel Stanovsky, Swabha Swayamdipta, Jesse Dodge, and Noah~A.
  Smith. 2020.
\newblock \href {https://doi.org/10.18653/v1/2020.acl-main.593} {The right tool
  for the job: Matching model and instance complexities}.
\newblock In \emph{Proceedings of the 58th Annual Meeting of the Association
  for Computational Linguistics}, pages 6640--6651, Online. Association for
  Computational Linguistics.

\bibitem[{Shen et~al.(2019)Shen, Dong, Ye, Ma, Yao, Gholami, Mahoney, and
  Keutzer}]{qbert}
Sheng Shen, Zhen Dong, Jiayu Ye, Linjian Ma, Zhewei Yao, Amir Gholami,
  Michael~W. Mahoney, and Kurt Keutzer. 2019.
\newblock \href {http://arxiv.org/abs/1909.05840} {{Q-BERT:} hessian based
  ultra low precision quantization of {BERT}}.
\newblock \emph{CoRR}, abs/1909.05840.

\bibitem[{Socher et~al.(2013)Socher, Perelygin, Wu, Chuang, Manning, Ng, and
  Potts}]{sst}
Richard Socher, Alex Perelygin, Jean Wu, Jason Chuang, Christopher~D. Manning,
  Andrew Ng, and Christopher Potts. 2013.
\newblock \href {https://aclanthology.org/D13-1170} {Recursive deep models for
  semantic compositionality over a sentiment treebank}.
\newblock In \emph{Proceedings of the 2013 Conference on Empirical Methods in
  Natural Language Processing}, pages 1631--1642, Seattle, Washington, USA.
  Association for Computational Linguistics.

\bibitem[{Sun et~al.(2022)Sun, Liu, Zhu, Geng, Wu, He, Ni, Xie, Huang, and
  Qiu}]{hashee}
Tianxiang Sun, Xiangyang Liu, Wei Zhu, Zhichao Geng, Lingling Wu, Yilong He,
  Yuan Ni, Guotong Xie, Xuanjing Huang, and Xipeng Qiu. 2022.
\newblock \href {http://arxiv.org/abs/2203.01670} {A simple hash-based early
  exiting approach for language understanding and generation}.

\bibitem[{Sun et~al.(2020)Sun, Yu, Song, Liu, Yang, and Zhou}]{mobile_bert}
Zhiqing Sun, Hongkun Yu, Xiaodan Song, Renjie Liu, Yiming Yang, and Denny Zhou.
  2020.
\newblock \href {https://doi.org/10.18653/v1/2020.acl-main.195}
  {{M}obile{BERT}: a compact task-agnostic {BERT} for resource-limited
  devices}.
\newblock In \emph{Proceedings of the 58th Annual Meeting of the Association
  for Computational Linguistics}, pages 2158--2170, Online. Association for
  Computational Linguistics.

\bibitem[{Teerapittayanon et~al.(2017)Teerapittayanon, McDanel, and
  Kung}]{branchynet}
Surat Teerapittayanon, Bradley McDanel, and H.~T. Kung. 2017.
\newblock \href {http://arxiv.org/abs/1709.01686} {Branchynet: Fast inference
  via early exiting from deep neural networks}.
\newblock \emph{CoRR}, abs/1709.01686.

\bibitem[{Vaswani et~al.(2017)Vaswani, Shazeer, Parmar, Uszkoreit, Jones,
  Gomez, Kaiser, and Polosukhin}]{transformer}
Ashish Vaswani, Noam Shazeer, Niki Parmar, Jakob Uszkoreit, Llion Jones,
  Aidan~N Gomez, \L~ukasz Kaiser, and Illia Polosukhin. 2017.
\newblock \href
  {https://proceedings.neurips.cc/paper_files/paper/2017/file/3f5ee243547dee91fbd053c1c4a845aa-Paper.pdf}
  {Attention is all you need}.
\newblock In \emph{Advances in Neural Information Processing Systems},
  volume~30. Curran Associates, Inc.

\bibitem[{Voita et~al.(2019)Voita, Talbot, Moiseev, Sennrich, and
  Titov}]{prune_heads}
Elena Voita, David Talbot, Fedor Moiseev, Rico Sennrich, and Ivan Titov. 2019.
\newblock \href {http://arxiv.org/abs/1905.09418} {Analyzing multi-head
  self-attention: Specialized heads do the heavy lifting, the rest can be
  pruned}.

\bibitem[{Wang et~al.(2018)Wang, Singh, Michael, Hill, Levy, and Bowman}]{glue}
Alex Wang, Amanpreet Singh, Julian Michael, Felix Hill, Omer Levy, and
  Samuel~R. Bowman. 2018.
\newblock \href {http://arxiv.org/abs/1804.07461} {{GLUE:} {A} multi-task
  benchmark and analysis platform for natural language understanding}.
\newblock \emph{CoRR}, abs/1804.07461.

\bibitem[{Wang et~al.(2017)Wang, Hamza, and Florian}]{qqp}
Zhiguo Wang, Wael Hamza, and Radu Florian. 2017.
\newblock \href {http://arxiv.org/abs/1702.03814} {Bilateral multi-perspective
  matching for natural language sentences}.
\newblock \emph{CoRR}, abs/1702.03814.

\bibitem[{Warstadt et~al.(2019)Warstadt, Singh, and Bowman}]{cola}
Alex Warstadt, Amanpreet Singh, and Samuel~R. Bowman. 2019.
\newblock \href {https://doi.org/10.1162/tacl_a_00290} {Neural network
  acceptability judgments}.
\newblock \emph{Transactions of the Association for Computational Linguistics},
  7:625--641.

\bibitem[{Williams et~al.(2018)Williams, Nangia, and Bowman}]{mnli}
Adina Williams, Nikita Nangia, and Samuel Bowman. 2018.
\newblock \href {https://doi.org/10.18653/v1/N18-1101} {A broad-coverage
  challenge corpus for sentence understanding through inference}.
\newblock In \emph{Proceedings of the 2018 Conference of the North {A}merican
  Chapter of the Association for Computational Linguistics: Human Language
  Technologies, Volume 1 (Long Papers)}, pages 1112--1122, New Orleans,
  Louisiana. Association for Computational Linguistics.

\bibitem[{Xia et~al.(2022)Xia, Zhong, and Chen}]{cofi}
Mengzhou Xia, Zexuan Zhong, and Danqi Chen. 2022.
\newblock \href {http://arxiv.org/abs/2204.00408} {Structured pruning learns
  compact and accurate models}.

\bibitem[{Xin et~al.(2020)Xin, Tang, Lee, Yu, and Lin}]{deebert}
Ji~Xin, Raphael Tang, Jaejun Lee, Yaoliang Yu, and Jimmy Lin. 2020.
\newblock \href {https://doi.org/10.18653/v1/2020.acl-main.204} {{D}ee{BERT}:
  Dynamic early exiting for accelerating {BERT} inference}.
\newblock In \emph{Proceedings of the 58th Annual Meeting of the Association
  for Computational Linguistics}, pages 2246--2251, Online. Association for
  Computational Linguistics.

\bibitem[{Xin et~al.(2021)Xin, Tang, Yu, and Lin}]{berxit}
Ji~Xin, Raphael Tang, Yaoliang Yu, and Jimmy Lin. 2021.
\newblock \href {https://doi.org/10.18653/v1/2021.eacl-main.8} {{BER}xi{T}:
  Early exiting for {BERT} with better fine-tuning and extension to
  regression}.
\newblock In \emph{Proceedings of the 16th Conference of the European Chapter
  of the Association for Computational Linguistics: Main Volume}, pages
  91--104, Online. Association for Computational Linguistics.

\bibitem[{Zhou et~al.(2020)Zhou, Xu, Ge, McAuley, Xu, and Wei}]{pabee}
Wangchunshu Zhou, Canwen Xu, Tao Ge, Julian McAuley, Ke~Xu, and Furu Wei. 2020.
\newblock \href
  {https://proceedings.neurips.cc/paper_files/paper/2020/file/d4dd111a4fd973394238aca5c05bebe3-Paper.pdf}
  {Bert loses patience: Fast and robust inference with early exit}.
\newblock In \emph{Advances in Neural Information Processing Systems},
  volume~33, pages 18330--18341. Curran Associates, Inc.

\end{thebibliography}
\bibliographystyle{acl_natbib}

\newpage
\appendix
\section{Background and Related Work}
\paragraph{Efficient non-adaptive inference} In the context of non-adaptive methods, numerous studies in the literature have improved the inference efficiency of transformers through various techniques such as knowledge distillation \cite{knowledge_distillation} into smaller models \cite{distilbert,mobile_bert,dynabert,tinybert}, pruning unimportant weights of the model \cite{cofi,block_pruning,movement_pruning,metapruning}, and weight quantization \cite{qbert,ibert} to store weights of a network with lower precision values. The aforementioned approaches result in smaller and more efficient albeit fixed and static models.
\paragraph{Efficient adaptive inference.} Another line of work have proposed adaptive inference methods that allow the network to allocate varying amounts of compute for each sample. The predominant technique in this area is early exiting via adding internal classifiers to intermediate layers \cite{RTJ,shallow_deep,deebert,branchynet,pabee,fastbert}: 
To early exit, prior work either use the confidence score of internal classifiers' predictions \cite{RTJ,deebert}; the entropy of these predictions \cite{fastbert}; a module that predicts whether a layer should early exit or not \cite{berxit}; a patience based change in these internal prediction \cite{pabee}; or a hash based mechanism to do token level early exiting \cite{hashee}.

\citet{devvrit2023matformer} recently proposed MatFormer that can enable adaptive compute based on the resource constraints but does not utilize dynamic token-based routing making \method complementary to it.

\section{Implementation Details}
\label{impl_details_appendix}
\subsubsection{Transformer Networks} Transformer networks \cite{transformer} are composed of a stack of $L$ layers and each layer has two main components: Multi-head attention (MHA) and Feed-forward network (FFN).

\paragraph{MHA} consists of $n_h$ heads where each head computes the attention operation \cite{transformer} on the projection of a sequence of $l$ tokens $x = (x_1, x_2, ..., x_l)$$ $ into key, query, and values: $o^{i}_{\text{head}} = \text{Attention}(\textbf{W}_Kx, \mathbf{W_Q}x, \mathbf{W_V}x)$,
wherein  $1\!\leq\!i\!\leq\!H$, $o^{i}_{\text{head}} \in \mathbb{R}^{d_{\text{head}}}$ and $\mathbf{W}_K, \mathbf{W}_Q, \mathbf{W}_V \in \mathbb{R}^{d_{\text{model}} \times d_{\text{head}}}$ are the key, query, and value projection matrices, and $d_\text{head} = d_\text{model} / n_h$. The outputs from different heads are concatenated into $o_{\text{heads}} \in \mathbb{R}^{d_{\text{model}}}$ and projected with another matrix $\mathbf{W}_O \in \mathbb{R}^{d_{\text{model}} \times d_{\text{model}}}$ and passed through a layer norm \cite{layernorm} to get the output: $o_{\text{MHA}} = \text{LayerNorm} (x + \textbf{W}_oo_{\text{heads}}).$

\paragraph{FFN} consists of two feed-forward layers $W_1$ and $W_2$ with $\text{GeLU}$ \cite{gelu} non-linearity and takes the output of MHA module and computes $\text{LayerNorm}(o_{\text{MHA}} + \text{GeLU}(\textbf{W}_1o_{\text{MHA}})\textbf{W}_2).$

\subsubsection{Reducing Width of a Transformer Layer}
\label{sub_adaptive_component}

\begin{figure}[h!]
    \raggedright
    \includegraphics[width=\linewidth]{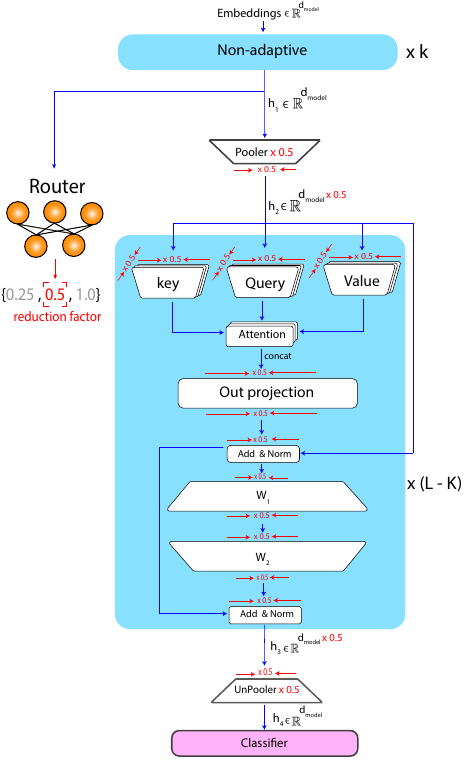}
    \caption{Detailed schematic of reducing with of different components in a transformer model based on the prediction of the router.}
    \label{fig:detailed_width_reduction}
\end{figure}

Figure \ref{fig:detailed_width_reduction} illustrates our approach in reducing the capacity of a transformer network based on the given reduction factor.
We leave the first $K$ layers (aka non-adaptive module) of the model intact, where $K$ is a hyperparameter. Given an input sequence $x$, after computing the intermediate representation $h \in \mathbb{R}^{l \times d_{\text{model}}}$ by the non-adaptive module, and the reduction factor $r$ by the router, we reduce the width of different components which we will describe in detail next.

Before passing $h$ to adaptive layers, we pass it through a pooler module that reduces its dimensionality to $d_{\text{model}} \times r$. For example, if $d_{model}=768$ and $r=0.25$, the input to the adaptive module will be 192 dimensional. Although we experimented with affine transformations for the pooler, in practice we observed that a simpler pooling operation such as selecting just the first  $d_{model} \times r$ dimensions works as good. Therefore, as this pooler is simpler and importantly adds no extra FLOPs to inferece, we opted for that. For the unpooler, we used an affine transformation that transforms the hidden state with reduced dimensionality (i.e. $r \cdot d_{\text{model}}$) to $d_{model}$. Note that for the unpooler layer, the dimensionality of the input can change while the output dimensionality if always $d_{\text{model}}$. 

\begin{figure*}[ht!]
     \centering
     \begin{subfigure}{0.4\textwidth}
         \centering
         \includegraphics[width=\textwidth]{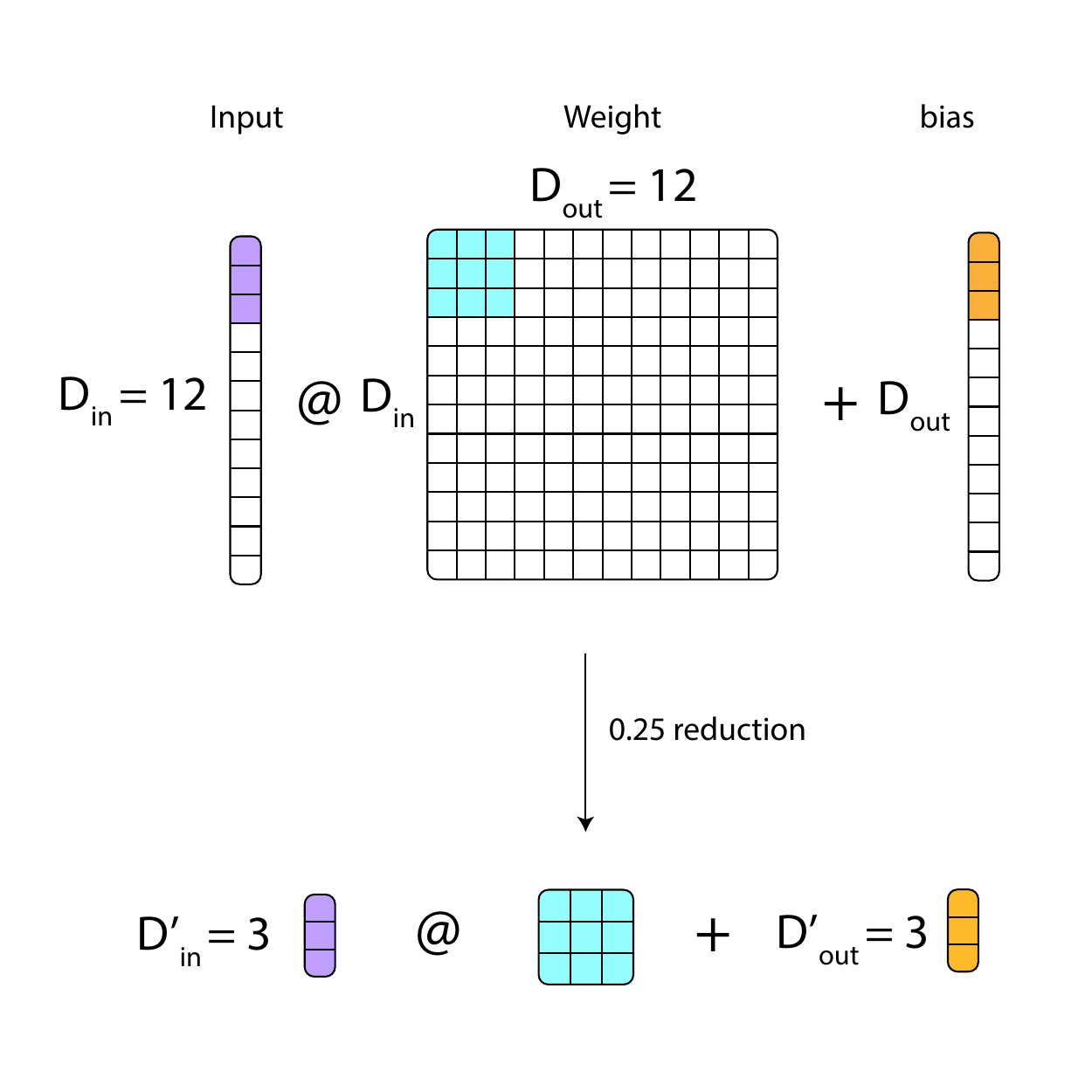}
         \caption{FFN}
         \label{fig:y equals x}
     \end{subfigure}
     \begin{subfigure}{0.4\textwidth}
         \centering
         \includegraphics[width=\textwidth]{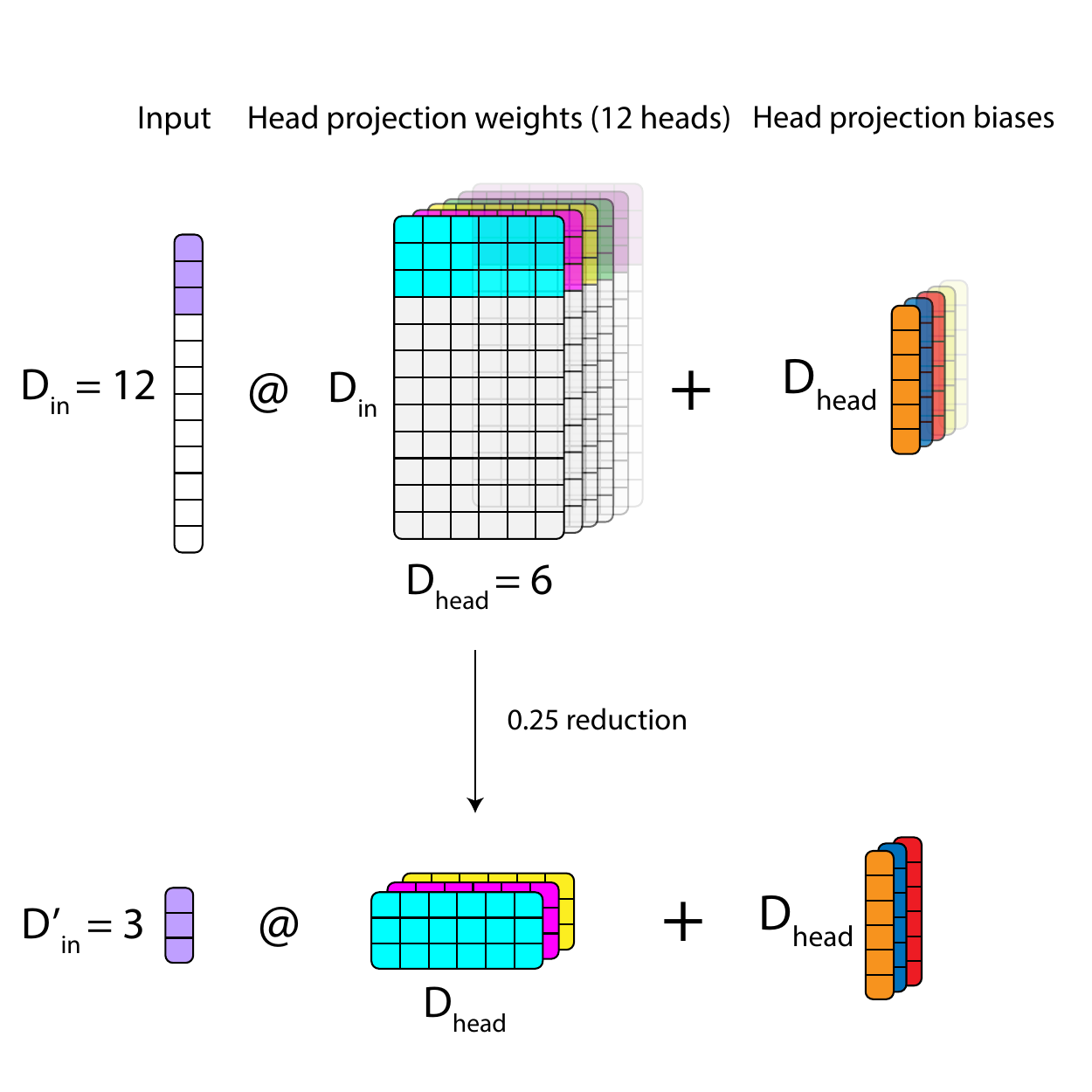}
         \caption{MHA}
         \label{fig:three sin x}
     \end{subfigure}
     \hfill
    \caption{Reducing the width of FFN and MHA in adaptive layers.}
    \label{fig:head_ff_dim_reduction}
\end{figure*}

Throughout this paper, whenever we decrease a dimension of a component (e.g. input vectors or weights) from $d$ to $d^\prime$, we exclusively use the first $d'$ dimensions and disregards the remainder in the computations. This is illustrated schematically in Figure \ref{fig:head_ff_dim_reduction} for FFN and MHA components in transformer models.

In what follows we describe how we reduce the width of different components 
by the reduction factor $r$ which is also depicted in detail in Figure \ref{fig:detailed_width_reduction}. 
\paragraph{Reducing MHA Width.} We reduce the width of MHA 
using the following steps:

1. The input dimension $d_{\text{model}}$ of self-attention projection matrices $\mathbf{W}_K$, $\mathbf{W}_Q$, and $\mathbf{W}_v$ is decreased to $d_{\text{model}} \cdot r$. We do not change the output dimension $d_{\text{head}}$ of the linear projections.

2.
 The adaptive module only computes the output of the first $n_h \cdot r$ heads and disregards the other heads. Therefore the dimensionality of $o_{\text{heads}}$ is decreased to $d_{\text{model}} \cdot r$. Note that we could also reduce the dimensionality of each head instead, however, we built on previous findings in the literature that many of the heads in transformers are redundant and can be pruned without hurting the performance \cite{remove_heads,prune_heads}.

3. For the output projection $\mathbf{W}_o$ of MHA, the input and output dimensions, which are both equal to $d_{\text{model}}$, are reduced to $d_{\text{model}} \cdot r$.

Throughout all the changes above, we do not alter the sequence length of the input (i.e. $l$) or any of the hidden states. 
\paragraph{Reducing FFN width.} Similar to the output projection $\mathbf{W}_o$ of MHA, the input and output dimensions of the feed-forward layers $\mathbf{W}1$ and $\mathbf{W}2$ in the FFN are reduced by a factor of $1/r$. Therefore, the output dimension of all the adaptive layers
  are reduced by a factor of $r$ to $D_{\text{model}} \times r$.

It is important to note that majority of operations in FFN and MHA components are comprised of matrix multiplications and reducing the input and output dimensions of these multiplications by a factor of $r_j$ will reduce their flops by a factor of $r_j^2$. 
\paragraph{Reducing layernorms width.} We also reduce the width of the layer norm parameters by a factor of $r$. In our experiments, we find it beneficial to initialize the layernorm parameters with the corresponding first dimensions of original layernorm parameters instead of training them from scratch.

\subsubsection{Inference}
\label{section_inference}
In contrast to training where the width of the adaptive layers doing forward pass is enforced by the router labels, at inference, the router predicts the reduction factor. More formally, given the input sample $x$ and router logits $\hat{W} \in \mathbb{R}^M$, we select the reduction factor $r_j$ to do the forward pass where j is $\argmax(\hat{W})$.

\section{Training Details}
\label{training_details}
Similar to \cite{dynabert}, we reorder the heads based on their importance before fine-tuning on the down-stream task. We set $\lambda_{task}$ to 1 and $\lambda_{router}$ to 0.5 in our approach and use a set of two reduction factors $\{0.25,1.0\}$. We choose the batch size in $\{16, 32\}$ depending on the model and dataset. We do a grid search over the lower and upper bound on the confidence thresholds and do the forward pass of each reduction factor on a separate GPU. We choose the window size values in $\{3, 5\}$. We set $\lambda_{task}$ to 1 and $\lambda_{router}$ to 0.5 in our approach and use a set of two reduction factors $\{0.25,1.0\}$. We do a grid search over the lower and upper bound on the confidence thresholds and the hyperparamters of the baselines and do the forward pass of each reduction factor on a separate GPU. Similar to previous work \cite{dynabert}, we pad the input sequences in a batch to 128 tokens while training.
To train \method with Transkimer on RoBERTa, we first train RoBERTa + Transkimmer with skim coefficient of $0.3$. We then start from this model as an initial checkpoint and train for 10 epochs using both our loss and Transkimmer loss. Here, training the router is done with with $0.25$ and $1.0$ as reduction factors.



\section{Results Details}
\subsubsection{GLUE Results} 
\label{glue_full_results}
We do a thorough hyperparameter search with different sample adaptive inference methods to get the accuracy of the RoBERTa\textsubscript{base} model across different FLOPs values. We report the total number of FLOPs on the validation set and do not pad the input sequences before feeding to the model. Figure \ref{full_glue_results} shows the plots for 8 of the sub-tasks in the GLUE benchmark.

\subsubsection{Different Backbone Results} 
\label{different_models_full_results}
Table \ref{full_table_different_models} shows the comparison of different sample adaptive inference methods applied to different backbones.

\tablemodelsappendix

\section{Ablations and Discussion}
\label{appendix_ablation_discussion}

\subsubsection{Router Ablation}
\begin{figure}[h!]
    \centering
    \includegraphics[width=\linewidth]{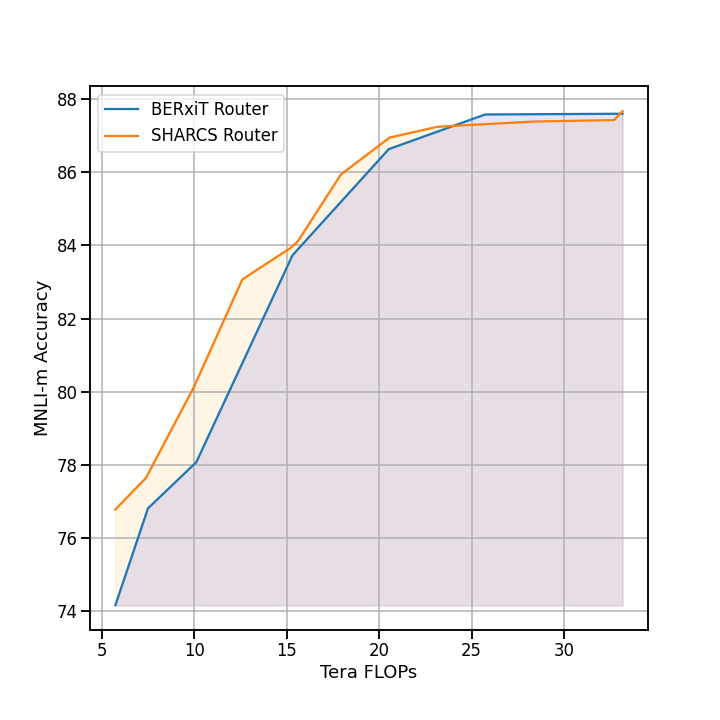}
    \caption{Accuracy vs. FLOPs plot for Adaptive width sub-networks trained with \method router (orange) and BERxiT \cite{berxit} router (blue).}
    \label{router_ablation_auc}
\end{figure}

Figure \ref{router_ablation_auc} shows the accuracy FLOPs plot for two methods: 1) The router introduced in \method with adaptive width sub-networks. 2) A router similar to the early-exiting router used in BERxiT \cite{berxit} with adaptive-width sub-networks. We report the AUC for a similar range of FLOPs for both methods and scale the accuracy and FLOPs values to $[0, 1]$.

\subsubsection{Number of Reduction Factors}
\begin{figure}[h!]
    \centering
    \includegraphics[width=0.8\linewidth]{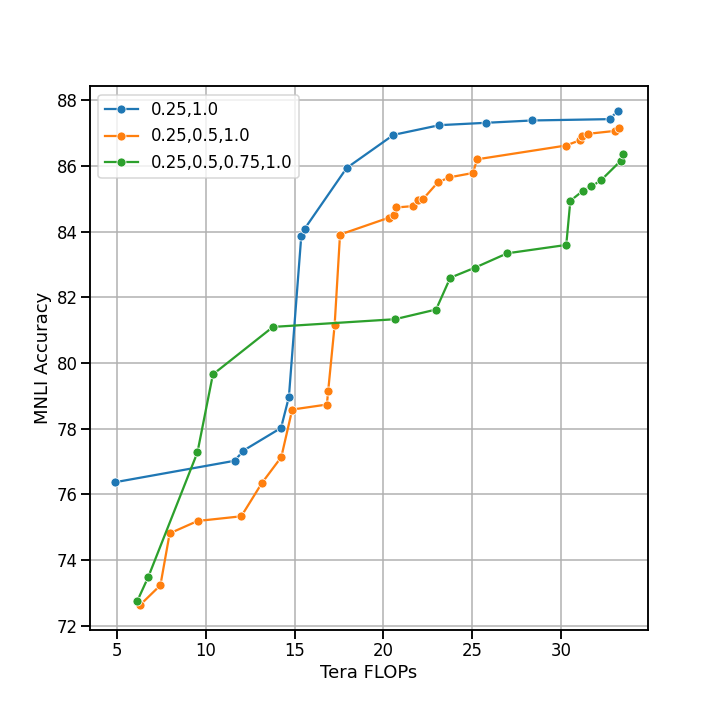}
    \caption{Results of different number of sub-networks (or reduction factors) with RoBERTa\textsubscript{base} model on MNLI-m.}
    \label{fig:ablation_n_subnetworks}
\end{figure}

We change the number of reduction factors (or hardness levels) $M$ to three and four and use the reduction factors $\{0.25,0.5,1.0\}$ and $\{0.25,0.5,0.75,1.0\}$ respectively. Figure \ref{fig:ablation_n_subnetworks} shows the results of changing the number of reduction factors with RoBERTa\textsubscript{base} model on MNLI-m dataset. The model trained with two reduction factors outperform the other cases for FLOPs values above 15 teraFLOPs significantly and four reduction factors can get better results in the range of 10-15 Tera FLOPs. 

\subsubsection{Confidence Thresholds}
\begin{figure}[h!]
    \centering
    \includegraphics[width=0.8\linewidth]{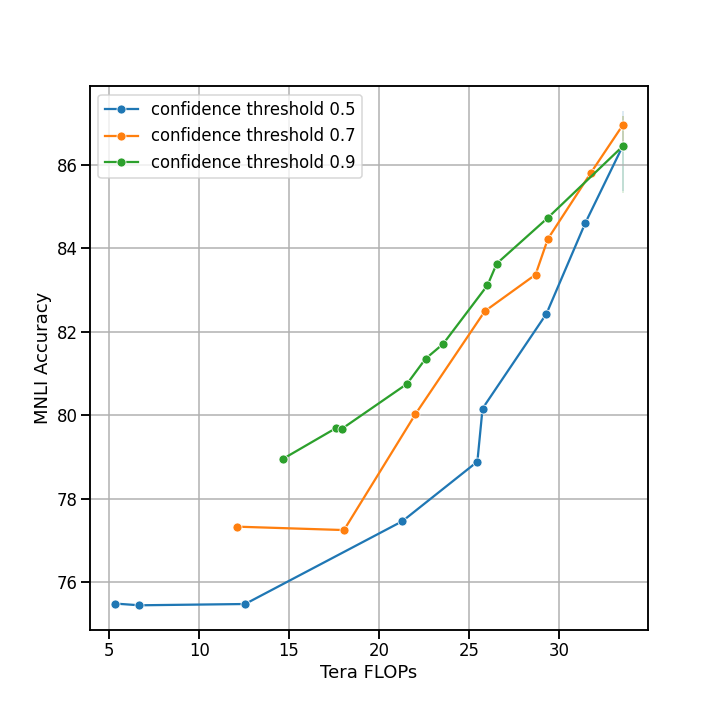}
    \caption{Results of changing confidence thresholds with RoBERTa\textsubscript{base} model on MNLI-m dataset. We use reduction factors $\{0.25,1.0\}$ and confidence thresholds $[0, x]$ for reduction factor $0.25$ and confidence thresholds $[x,1]$ for reduction factor 1, where $x \in \{0.5, 0.7. 0.9\}$.}
    \label{fig:ablation_conf_thresholds}
\end{figure}

We study the effect of changing confidence thresholds on our results with a simple experiment: With two reduction factors $\{0.25, 1.0\}$ we use the following confidence score lower and upper bounds: $[0.0, x]$ for full network and $[x, 1.0]$ for 0.25 network, where $x \in \{0.5, 0.7, 0.9\}$. We place the router after layer 1 (i.e. $K=1$). 
Figure \ref{fig:ablation_conf_thresholds} shows the results of RoBERTa\textsubscript{base} model on MNLI-m dataset. According to the figure, higher value of $x$ leads to better acuracies for a fixed number of FLOPs. Furthermore, as we decrease $x$, we can reach lower values of FLOPs. This is intuitive as decreasing the lower bound of smallest reduction factor makes more samples be routed to that which reduces the overall FLOPs.

\begin{figure}[h!]
    \centering
    \includegraphics[width=\linewidth]{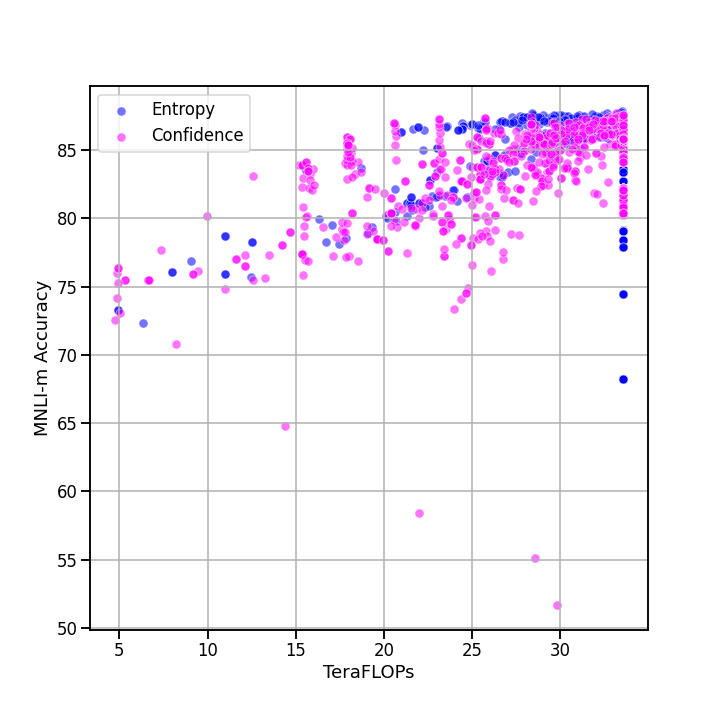}
    \caption{Comparing entropy and confidence based hardness labels for training the router.}
    \label{entropy_ablation}
\end{figure}

\begin{figure}[t!]
    \centering
    \includegraphics[width=0.7\linewidth]{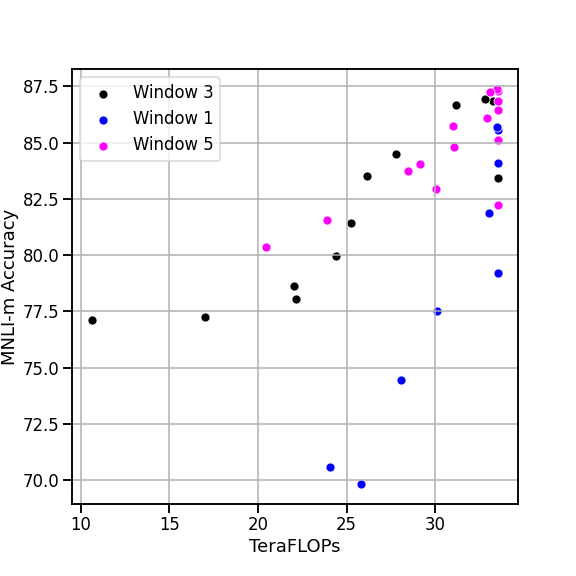}
    \caption{Effect of changing history window size in \method on MNLI-m Accuracy FLOPs trade-off.}
    \label{ablate_window_size}
\end{figure}

\subsubsection{Using Entropy or Confidence-based Hardness Labels to Train Router} Similar to sample adaptive inference in early exiting methods \cite{deebert,fastbert}, one can get hardness labels by defining thresholds on entropy of the network predictions instead of the confidence. We use the following formula to compute the entropy of network's prediction:
$$H = -\sum_{i=1}^{C}p_i\cdot\log{p_i},$$
wherein $C$ denotes the number of classes of the classifier, and $p_i$ is the softamx probablity that the classifier assigns to class $i$.
Figure \ref{entropy_ablation} shows the accuracy FLOPs trade-off for both of these metrics on MNLI-m dataset with RoBERTa\textsubscript{base} model. Given that the confidence score is a simpler approach and does not require any additional computation, we opted for using that in our method.

\subsubsection{Changing History Window Size in Training the Router} 
\label{appendix_history_window}
As mentioned in (\sectionref{section_hardness_label}), having a larger window size helps stabilizing training the router as the hardness label for a training sample might change throughout training. To illustrate this effect, we train \method on RoBERTa\textsubscript{base} model with three different window sizes $\{1, 3, 5\}$ for 10 epochs on MNLI-m dataset. We place the router at layer 2, use two reduction factors $\{0.25, 1.0\}$, and set the confidence thresholds to $[0.0, 0.8]$ for reduction factor 1 and $[0.8, 1.0]$ for reduction factor 0.25.  Figure \ref{ablate_window_size} shows the accuracy FLOPs trade offs that different checkpoints with different window sizes get throughout training on the validation set. According to the figure, the network can reach higher accuracy when trained with window size 3 or 5. We did not see any significant improvements by using a window size larger than 5.


\begin{figure}[t!]
    \centering
    \includegraphics[width=0.9\linewidth]{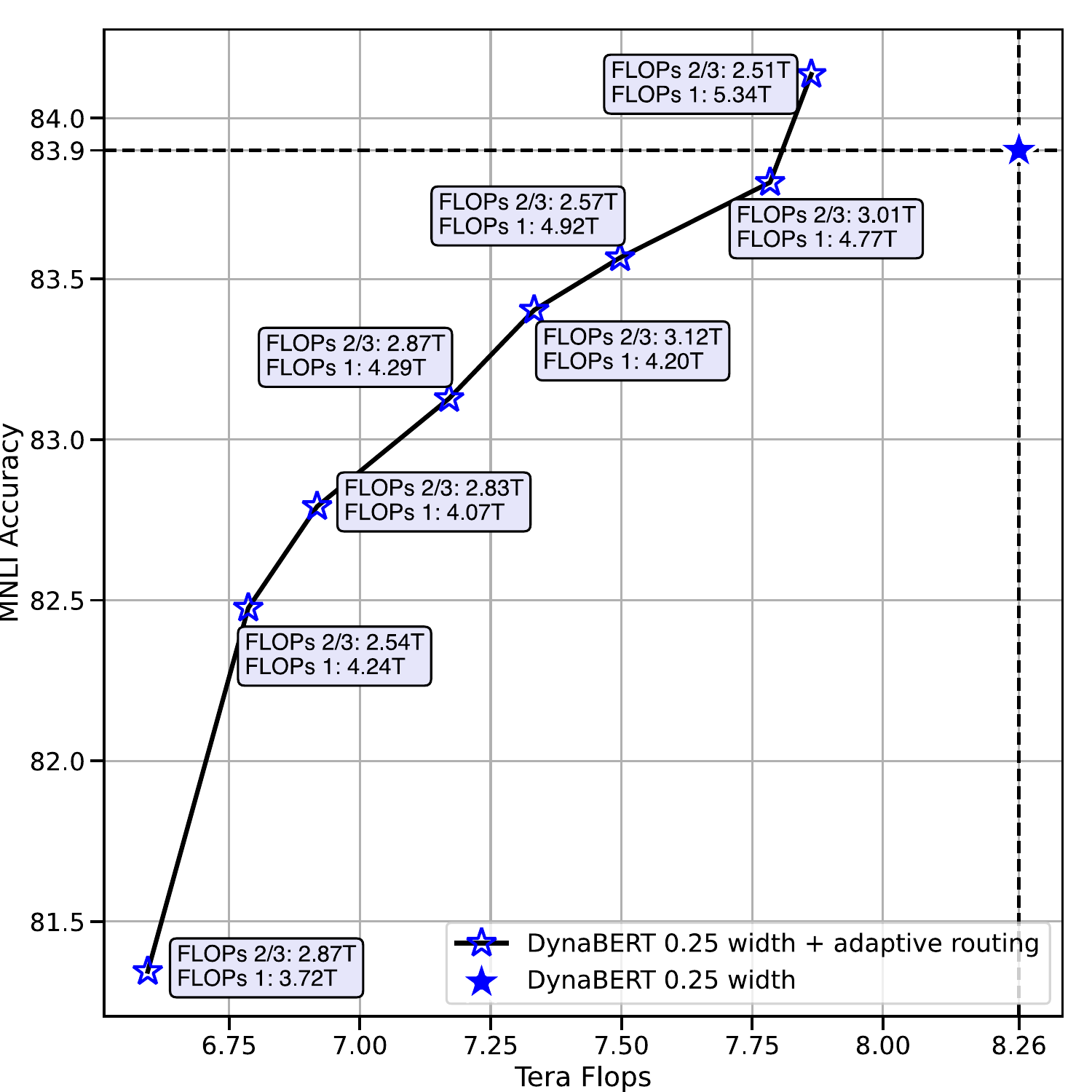}
    \caption{Accuracy FLOPs plot for \method applied to DynaBERT 0.25 W \cite{dynabert} model. We set the reduction factors to  $\{2/3, 1.0\}$. By placing router at different layers and changing the confidence thresholds, we can get different points in the plot. Note that the FLOPs for each sub-network in each point is also reported.}
    \label{fig:dyna_acc_flops_plot}
\end{figure}

\begin{figure}[t!]
    \centering
    \includegraphics[width=0.9\linewidth]{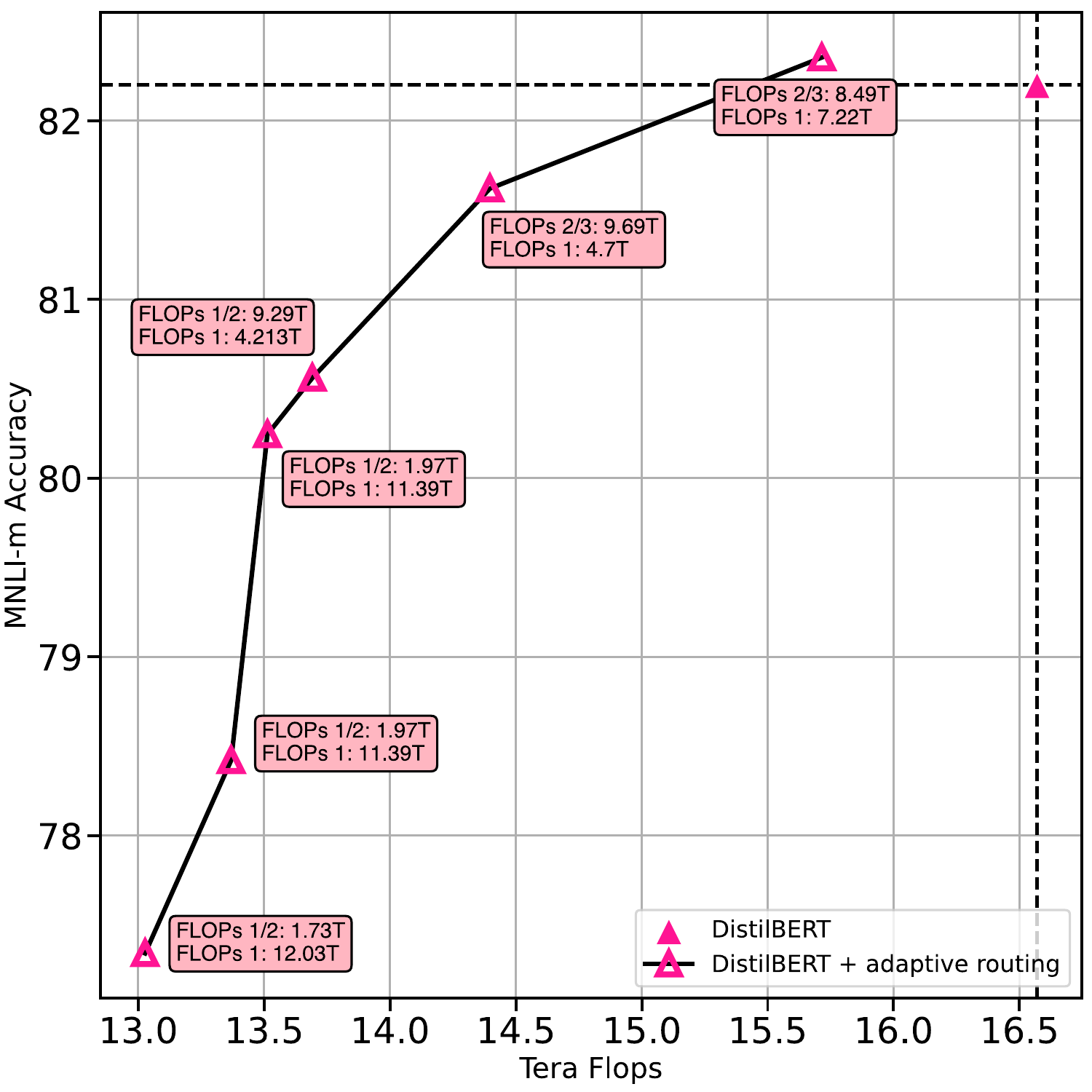}
    \caption{Accuracy FLOPs plot for \method applied to DistilBERT\cite{distilbert} model. We set the reduction factors to  $\{2/3, 1.0\}$ or $\{0.5, 1.0\}$. By placing router at different layers and changing the confidence thresholds, we can get different points in the plot. Note that the FLOPs for each sub-network in each point is also reported.}
    \label{fig:distilbert_acc_flops_plot}
\end{figure} 
\begin{figure*}[ht!]
    \raggedright
    \includegraphics[width=\textwidth]{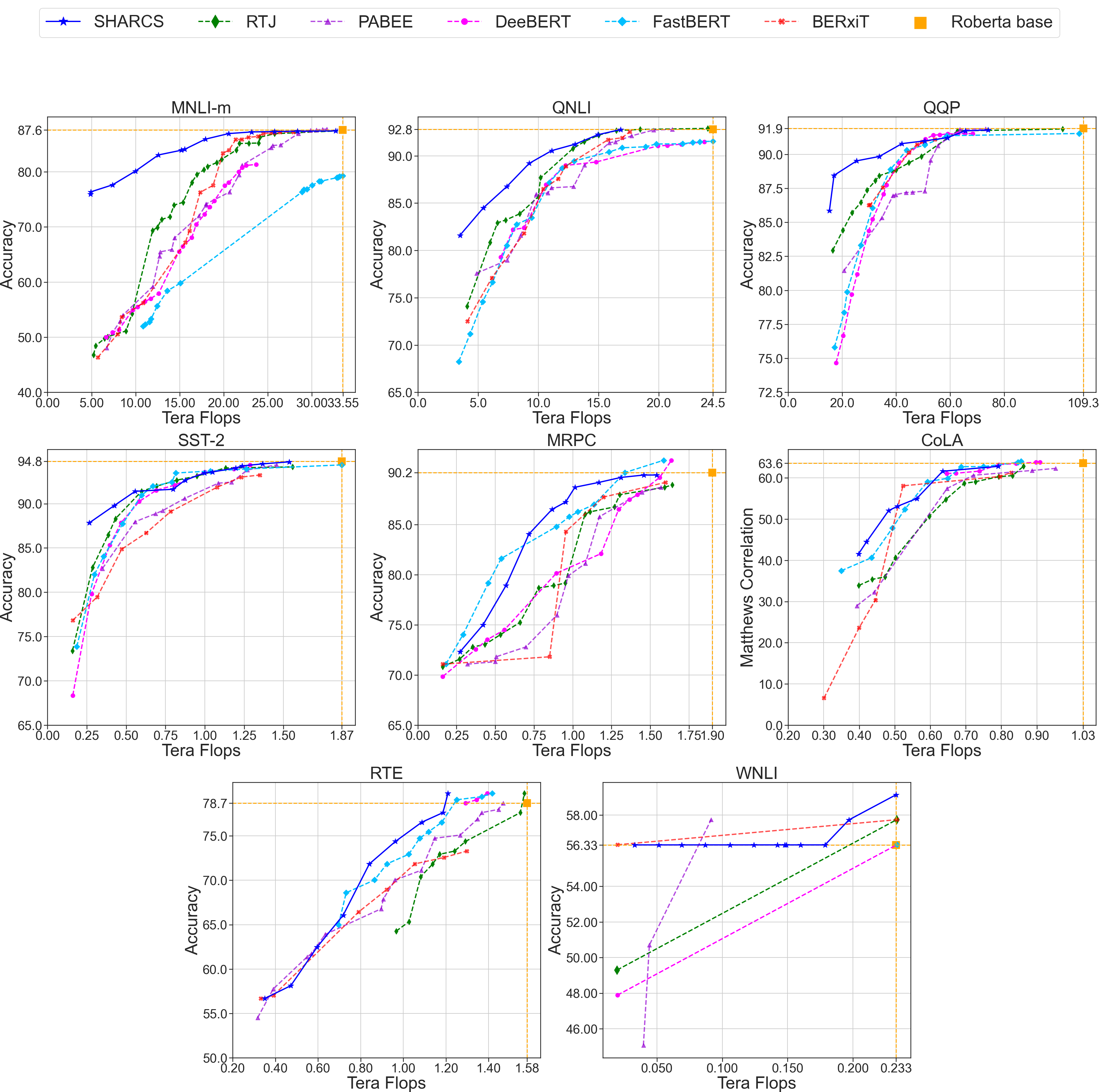}
    \caption{GLUE benchmark results. Best viewed in color.}
    \label{full_glue_results}
\end{figure*}

\end{document}